\documentclass[letterpaper, 10 pt, conference]{ieeeconf}  

\usepackage{amsmath}
\usepackage{amssymb}
\usepackage{lipsum}
\usepackage{booktabs}
\usepackage{graphicx}
\usepackage{array}
\usepackage{tabularx}
\usepackage{color}
\usepackage{tcolorbox}
\tcbuselibrary{listings,breakable}
\usepackage{listings}
\usepackage{multirow}

\usepackage[backend=biber,style=ieee,citestyle=numeric-comp,sorting=none]{biblatex}
\IEEEoverridecommandlockouts                           

\title{\LARGE \bf
Coordinated Control of Multiple Construction Machines Using LLM-Generated Behavior Trees with Flag-Based Synchronization
}

\author{
  Akinosuke Tsutsumi$^{1}$, Tomoya Itsuka$^{1}$, Yuichiro Kasahara$^{1}$, Tomoya Kouno$^{1}$, Kota Akinari$^{1}$,\\
  Genki Yamauchi$^{2}$, Daisuke Endo$^{2}$, Taro Abe$^{2}$, Takeshi Hashimoto$^{2}$,\\
  Keiji Nagatani$^{3}$, and Ryo Kurazume$^{4}$%
  \thanks{$^{1}$Graduate School of Information Science and Electrical Engineering,
  Kyushu University, Fukuoka, Japan
  \{tsutsumi,itsuka,kasahara,kouno,akinari\}@irvs.ait.kyushu-u.ac.jp}%
  \thanks{$^{2}$Public Works Research Institute, Ibaraki, Japan
  \{yamauchi-g573bs,endou-d177cl,abe-t575co,t-hashimoto\}@pwri.go.jp}%
  \thanks{$^{3}$Institute of Systems and Information Engineering,
  University of Tsukuba, Ibaraki, Japan
  nagatani@cs.tsukuba.ac.jp}%
  \thanks{$^{4}$Faculty of Information Science and Electrical Engineering,
  Kyushu University, Fukuoka, Japan
  kurazume@ait.kyushu-u.ac.jp}%
}

\begin{document}

\maketitle
\thispagestyle{empty}
\pagestyle{empty}

\begin{abstract}

Earthwork operations face increasing demand, while workforce aging creates a growing need for automation.
ROS2-TMS for Construction, a Cyber-Physical System framework for construction machinery automation, has been proposed; however, its reliance on manually designed Behavior Trees (BTs) limits scalability in cooperative operations.
Recent advances in Large Language Models (LLMs) offer new opportunities for automated task planning, yet most existing studies remain limited to simple robotic systems.
This paper proposes an LLM-based workflow for automatic generation of BTs toward coordinated operation of construction machines.
The method introduces synchronization flags managed through a Global Blackboard, enabling multiple BTs to share execution states and represent inter-machine dependencies.
The workflow consists of Action Sequence generation and BTs generation using LLMs.
Simulation experiments on 30 construction instruction scenarios achieved up to 93\% success rate in coordinated multi-machine tasks.
Real-world experiments using an excavator and a dump truck further demonstrate successful cooperative execution, indicating the potential to reduce manual BTs design effort in construction automation.
These results highlight the feasibility of applying LLM-driven task planning to practical earthwork automation.

\end{abstract}

\section{INTRODUCTION}

Earthwork operations are facing increasing demand as aging social infrastructure requires large-scale maintenance and reconstruction.
However, such operations are increasingly avoided by younger generations due to their substantial physical demands and the inherent risks associated with hazardous working environments.
As a result, rapid workforce aging and the discontinuity in the transmission of specialized skills have become critical challenges in the civil engineering industry.
In this context, the integration of Artificial Intelligence (AI) and robotics technologies is expected to provide transformative solutions by enabling autonomous and coordinated operation of construction machinery.

Automation of construction machinery has been studied for decades~\cite{zhangAutonomousExcavatorSystem2021, chenExACTEndtoEndAutonomous2024, zhai2025ext}.
Among these initiatives, ROS2-TMS for Construction~\cite{ROS2TMSforConstructionKasahara2024} has been proposed as a Cyber-Physical System framework for orchestrating multiple construction machines.
The system is implemented on the Robot Operating System 2 (ROS 2)~\cite{ros2-kaiju} and integrates sensing, database management, task scheduling, motion planning, and low-level control to enable coordinated earthwork operations.
In this system, the task scheduler is implemented using Behavior Trees (BTs).
BTs provide a modular and hierarchical structure for representing task execution logic, organizing actions flows through nodes such as Sequence and Fallback to enable flexible and reactive task execution~\cite{iovino2022survey}.

\begin{figure}[t]
  \centering
  \includegraphics[width=0.9\columnwidth]{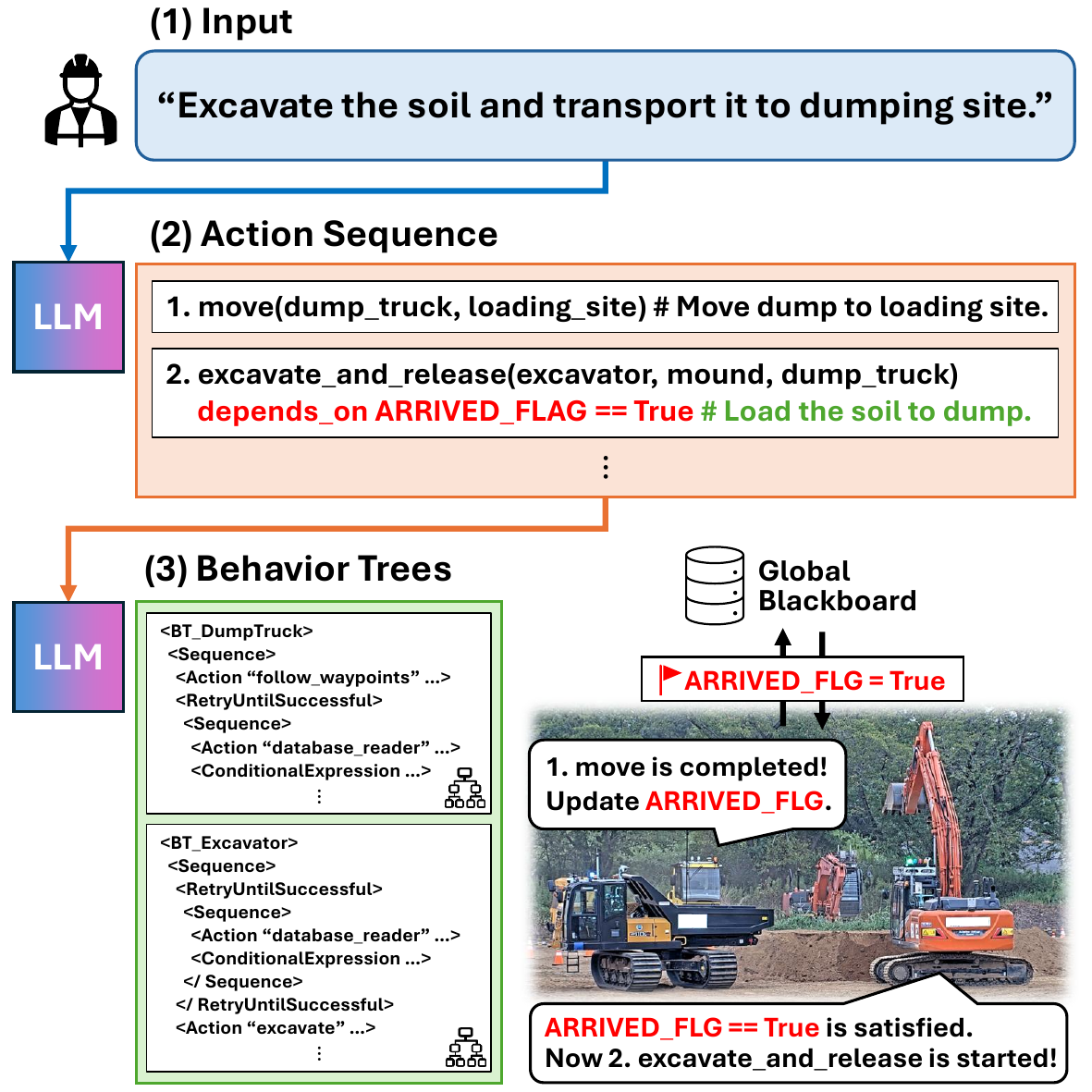}
  \caption{Overview of the proposed workflow for coordinated construction machine operation. 
           Natural language instruction is first converted into an Action Sequence by an LLM. 
           A second LLM then generates Behavior Trees for each machine. 
           Execution states are shared through a Global Blackboard, enabling flag-based synchronization between the excavator and dump trucks during operation.}
  \label{system_overview_simple}
  \vspace{-1em}
\end{figure}

While BTs are well suited for robotic control due to their modularity and reusability, they are manually designed by users in ROS2-TMS for Construction.
Designing BTs for single-machine tasks is manageable; however, cooperative operations involving multiple heterogeneous machines, such as excavators and dump trucks, introduce complex inter-machine dependencies and synchronization requirements.
As the scale and diversity of construction scenarios increase, manual BTs design becomes time-consuming, less scalable, and increasingly difficult to maintain.

Recent advances in Large Language Models (LLMs) have opened new possibilities for robotic task planning and BTs generation. Owing to their strong natural language understanding and reasoning capabilities, LLMs have been applied to instruction interpretation, symbolic world modeling using PDDL, and the generation of task structures~\cite{LLMBT2024, OBTEALLM2024}.
However, traditional symbolic task planning approaches struggle to scale to construction environments, where environmental information is highly complex and numerous objects and action constraints must be considered.

Some studies have further explored the direct generation of BTs from natural language instructions~\cite{BETRXPLLM2024, LLMasBTPlanner2025}.
Moreover, even approaches that rely entirely on LLMs for end-to-end BTs generation rarely address explicit coordination among heterogeneous machines or mechanisms for shared state management during execution~\cite{SMARTLLM2024, tian2025llm}.

To address these challenges, we propose an LLM-based workflow for automatic Behavior Tree generation for construction machinery automation.
The proposed method decomposes the generation process into two stages: (1) Action Sequence generation using an LLM, where task sequences and synchronization flags are produced, and (2) structured BTs generation based on predefined template structures.
To enable coordinated multi-machine execution, synchronization flags are managed through a shared Global Blackboard, allowing multiple BTs to exchange execution states and explicitly represent inter-machine dependencies. Furthermore, safety and feasibility are ensured by constraining the generation process using known machine-specific parameters stored in the system database. This design prevents invalid actions and reduces the risk of unsafe plans while preserving the flexibility of LLM-based reasoning.

By integrating LLM-based reasoning with a structured execution framework and shared state management, the proposed approach enables scalable and flexible task planning for heterogeneous construction machinery. The main contributions of this work are as follows:

\begin{itemize}
  \item An LLM-based two-stage workflow that first decomposes instructions into sequential tasks and then generates Behavior Trees.
  \item Coordinated control of multiple construction machines based on synchronization flags generated by LLM.
  \item Experimental validation demonstrating the effectiveness of the proposed method through simulation and real-world deployment on actual construction machinery.
\end{itemize}

\section{RELATED WORKS}

\subsection{Intelligent Construction Machinery}

Several studies have focused on excavation tasks using intelligent construction machinery. End-to-end autonomous control systems for excavators have been developed using imitation learning based on multimodal data acquired from LiDAR and cameras~\cite{zhangAutonomousExcavatorSystem2021, chenExACTEndtoEndAutonomous2024}. These systems integrate perception, motion planning, and control to enable autonomous excavation and material loading, and demonstrate long-duration autonomous operation under practical construction conditions.

More recently, ExT~\cite{zhai2025ext} has been proposed as a scalable autonomous excavation framework based on large-scale multi-task demonstration collection, transformer-based pretraining, and task-specific fine-tuning. The framework demonstrates improved generalization across diverse excavation conditions and sim-to-real transfer to physical excavator deployments.

Despite these advances, existing approaches primarily focus on single-excavator autonomy. However, earthwork operations inherently involve multiple types of construction machinery, such as excavators and dump trucks, whose coordination is essential for overall site productivity. Optimizing only a single machine does not necessarily improve the efficiency of the entire construction process.

Comprehensive automation of earthwork operations therefore requires coordination among heterogeneous machines. Such coordination introduces additional challenges, including task synchronization, shared state management, and dynamic scheduling, which remain insufficiently addressed in current intelligent excavation systems.

\subsection{Classical Task Planning for Robotics}

Classical approaches to robotic task planning are often formulated using symbolic representations such as STRIPS and PDDL~\cite{STRIPS, PDDLPlanningDomain}.

However, civil construction sites introduce additional complexities: (i) continuous and partially observable state spaces, (ii) dynamic changes in terrain and object configurations, and (iii) concurrent interactions among heterogeneous machines. Modeling such environments symbolically becomes increasingly difficult, and the large number of objects and actions significantly exacerbates the state explosion problem.

\begin{figure*}[t]
  \centering
  \includegraphics[width=\textwidth]{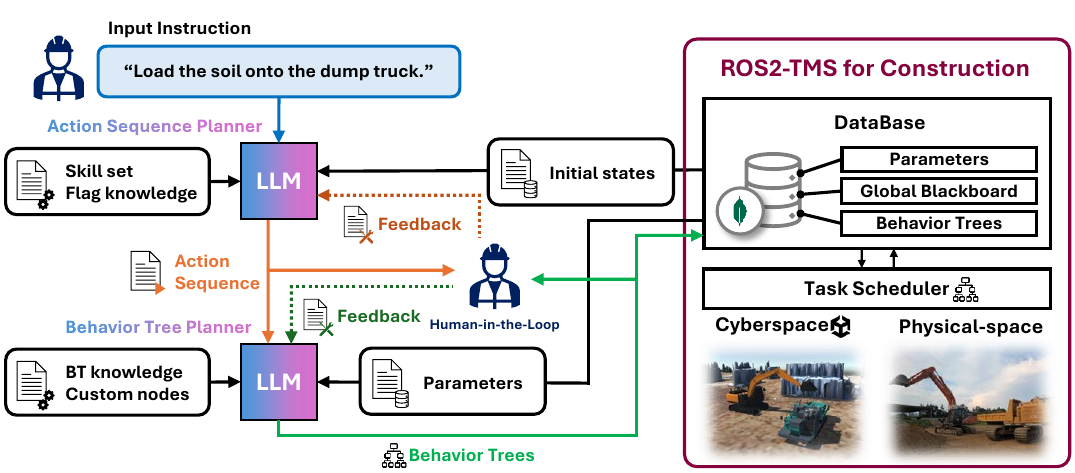}
  \caption{System architecture of the proposed framework. The LLM-based Action Sequence Planner decomposes natural language instructions into structured task representations, which are then transformed into executable BTs by the Behavior Tree Planner. Synchronization flags managed via a Global Blackboard enable coordinated multi-machine execution within a ROS2-TMS for Construction, with optional Human-in-the-Loop (HITL) refinement.}
  \label{fig:system_configuration}
  \vspace{-1em}
\end{figure*}

\subsection{LLM-based Task Planning and Behavior Tree Generation}

In recent years, robotic task planning using Large Language Models (LLMs) has attracted significant attention. 
Several studies have demonstrated that LLMs can effectively interpret natural language instructions and generate structured task plans. 
For example, SayCan~\cite{SayCan2022} enables LLMs to select appropriate actions from a predefined skill set by grounding them in real-world knowledge through pre-trained skills. 
ProgPrompt~\cite{ProgPrompt2022} generates executable robot programs directly from natural language instructions. 
SMART-LLM~\cite{SMARTLLM2024} further extends this direction to multi-robot settings by leveraging LLMs for task decomposition, coalition formation, and task allocation. 
These approaches highlight the capability of LLMs to perform high-level task reasoning.

More recently, the generation of Behavior Trees (BTs) has been explored. 
Some studies integrate LLMs with symbolic planning frameworks. 
LLM-BT~\cite{LLMBT2024} generates and parses PDDL-like structures from natural language prompts, while OBTEA~\cite{OBTEALLM2024} produces first-order logic formulas and performs condition tracing to guide planning. 
These approaches combine symbolic reasoning with LLM-based inference to enhance planning consistency.

Other studies focus on directly generating executable task structures. 
BETR-XP-LLM~\cite{BETRXPLLM2024} utilizes LLMs to address planning errors beyond classical planners while updating BTs construction policies to improve robustness. 
LLM-as-BT-Planner~\cite{LLMasBTPlanner2025} decomposes the process from intent understanding to BTs generation into multiple stages, enabling replanning through simulation feedback and human interaction. 
MRBTP~\cite{MRBTP2025a} applies LLMs to generate subtrees for heterogeneous multi-robot systems, improving BTs expansion efficiency. 
More recently, LLM-CBT~\cite{tian2025llm} proposes a closed-loop BTs generation framework for heterogeneous robot teams composed of unmanned ground vehicles (UGVs) and unmanned aerial vehicles (UAVs), introducing parallel nodes to enable concurrent behaviors and improve execution efficiency.

While these studies demonstrate the promise of LLM-based BTs generation for robotic systems, several limitations remain. 
Explicit synchronization mechanisms and shared state management across multiple BTs are rarely addressed, and most evaluations are conducted in simulation environments. 
Moreover, applications to large-scale heterogeneous construction machinery operating in dynamic earthwork environments remain largely unexplored.

\begin{table*}[t]
  \centering
  \vspace{2mm}
  \caption{List of available skills used for Action Sequence generation}
  \label{action_sequence_skills}

  \begin{tabularx}{\textwidth}{l l X}
  \toprule
  Machine Type & Skill & Description \\ 
  \midrule

  \textbf{Excavator} 
  & move(machine, destination) 
  & Move the specified machine to the destination. \\

  & initial\_pose(machine) 
  & Change the pose of the machine to its initial pose. \\

  & excavate\_and\_release(machine, excavate\_place, release\_place) 
  & Excavate soil from excavate\_place and release it to release\_place. \\

  & level(machine, level\_place) 
  & Level the soil at the specified level\_place.\\

  & gather(machine, gather\_place) 
  & Gather soil at the specified gather\_place.\\

  \midrule

  \textbf{Dump Truck}
  & move(machine, destination) 
  & Move the specified machine to the destination. \\

  & dump\_soil(machine) 
  & Dump the loaded soil at the current location. \\

  \bottomrule
  \end{tabularx}

\end{table*}

\begin{figure*}[t]
  \centering

  \begin{tcolorbox}[
  width=\textwidth,
  colback=white,
  colframe=black,
  boxrule=0.5pt,
  arc=1pt,
  left=3pt,
  right=3pt,
  top=3pt,
  bottom=3pt,
  boxsep=2pt
  ]

  \begin{lstlisting}[basicstyle=\ttfamily\scriptsize]
  1. initial_pose(excavator) # Return excavator to initial pose.
  2. move(dump_truck, loading_site) depends_on EXCAVATOR_INITIAL_POSE_FLG==true # Move dump truck next to excavator.
  3. excavate_and_release(excavator, mound, dump_truck) depends_on DUMPTRUCK_AT_LOADING_SITE_FLG==true and SENSING_ARRIVAL_FLG==true # Excavate and load the soil.

  EXCAVATOR_INITIAL_POSE_FLG: True when the excavator is in its initial pose; False otherwise.
  DUMPTRUCK_AT_LOADING_SITE_FLG: True when the dump truck is at the loading site; False otherwise.
  \end{lstlisting}

  \end{tcolorbox}
  
  \vspace{-1.5mm}
  \caption{Example of an Action Sequence generated for the instruction ``Load the soil onto a dump truck.''}
  \label{action_sequence_example}
  \vspace{-1em}

\end{figure*}

\section{PROPOSED METHOD}

Our proposed workflow for enabling coordinated operation between the construction machines is illustrated in Fig.~\ref{fig:system_configuration}.
The process of generating a Behavior Trees from user instructions consists of two main steps. 

In the first step, the LLM generates an Action Sequence based on the natural language instructions provided by the user. 
In the second step, the generated Action Sequence is transformed into a low-level Behavior Trees. 
The resulting BTs are then loaded into a task scheduler based on BehaviorTree.CPP~\cite{BehaviorTree_CPP} within the construction automation platform, ROS2-TMS for Construction, enabling execution either in simulation or on physical machines.

Inspired by prior work that decomposes planning into sequential stages of task structuring and BTs generation~\cite{LLMasBTPlanner2025}, we adopt a two-step workflow tailored to the specifications of ROS2-TMS for Construction. 
Unlike existing approaches, our method explicitly supports coordinated task planning for heterogeneous construction machines by introducing synchronization mechanisms that enable inter-machine dependencies during execution. 
The two steps of the proposed workflow are detailed in the following subsections.

\subsection{Action Sequence Generation}

An Action Sequence is a structured intermediate representation that specifies 
(i) an ordered set of abstract Skills assigned to construction machines and 
(ii) synchronization flags required to coordinate their execution. 
It serves as a bridge between instructions provided by the user and executable Behavior Trees.

The Action Sequence is generated by an LLM using the following inputs:
\begin{enumerate}
    \item Instructions provided by the user
    \item A predefined set of available \textit{skills}
    \item Predefined default flags and their initial states
    \item Rules for generating additional synchronization flags
\end{enumerate}
Default flags primarily represent environmental or system states, such as whether soil has been loaded into a dump truck (\texttt{SENSING\_LOADED\_FLG}), and are managed by the sensor system of the ROS2-TMS for Construction.

In addition, the Action Sequence generation process supports a Human-in-the-Loop (HITL) mechanism. 
After an initial Action Sequence is generated, users can review the proposed Skills and synchronization conditions. 
If there are issues such as inappropriate Skill selection or missing synchronization constraints, users may provide additional textual feedback to refine the instruction. 
The LLM then regenerates the Action Sequence based on the updated guidance.

\subsubsection*{Structure of the Action Sequence}

Formally, an Action Sequence $\mathcal{A}$ is defined as an ordered sequence
\[
\mathcal{A} = (A_1, A_2, \dots, A_n)
\]
where each $A_i$ denotes an \textit{Action Statement}.
Each $A_i$ is defined as follows:

\[
A_i = (S_i(M_i, \mathbf{p}_i), C_i, R_i)
\]

where:
\begin{itemize}
  \item $S_i$ denotes a Skill,
  \item $M_i$ denotes the construction machine executing the Skill,
  \item $\mathbf{p}_i$ denotes the parameters of the Skill,
  \item $C_i$ denotes a boolean precondition expressed over synchronization flags,
  \item $R_i$ denotes an optional reasoning comment.
\end{itemize}

The precondition $C_i$ is expressed as a boolean expression over flag-state pairs:

\[
C_i = \phi(F_1, F_2, \dots, F_n)
\]

where $\phi$ denotes a logical expression composed using conjunction ($\wedge$) and disjunction ($\vee$) operators.

\subsubsection*{Structure of the Action Sequence}

The Action Sequence is expressed in a Python-like format, as shown in Fig.~\ref{action_sequence_example}, and consists of the following components:

\begin{enumerate}

  \item \textit{Skill} —
  An abstract capability represented as a Python-like function,
  describing an intended operation assigned to a construction machine.
  The construction machine and task-related parameters,
  such as the destination or target pose, are explicitly specified as arguments.

  \item \textit{Precondition} —  
  A boolean condition that specifies the required state before a subsequent Skill can be considered.
  The required flag states are described using the \texttt{depends\_on} keyword.
  These conditions represent dependencies between Skills based on synchronization flags.

  \item \textit{Reasoning} —  
  A textual explanation associated with each Skill and its precondition,
  written as a Python-style comment (prefixed with “\#”).
  This explicit reasoning representation, inspired by Chain-of-Thought prompting~\cite{wei2022chain},
  encourages step-by-step decision making and improves the interpretability of the generated plan.

  \item \textit{Generated Flags} —  
  Additional synchronization flags created by the LLM beyond predefined default flags.
  When one construction machine must wait for another to complete a Skill,
  new flags are generated according to the following rules:
        \begin{itemize}
          \item The number of flags is not strictly limited, but redundant flags should be avoided when possible.
          \item Default flags are used to maintain synchronization of world states obtained through sensing systems; therefore, additional flags with similar semantic meanings may be generated to enable consistency checks between sensed states and machine operations.
          \item Flag names should clearly indicate the associated machine type and state.
          \item At the end of the Action Sequence, each generated flags are accompanied by a description specifying the conditions under which it becomes \texttt{True} or \texttt{False}.
        \end{itemize}

\end{enumerate}

This structured representation allows explicit modeling of inter-machine dependencies while maintaining compatibility with subsequent BTs generation.

\begin{table*}[t]
  \centering
  \vspace{2mm}
  \caption{Representative Construction Instruction Scenarios}
  \label{scenarios_example}
  \begin{tabularx}{\textwidth}{c X c c}
  \toprule
  No. & Instruction & Excavator & Dump Truck \\
  \midrule

  1 & \scriptsize Excavate once and load the soil at the temporary site. 
  & 1 & 0 \\

  2 & \scriptsize Perform three excavation actions with the excavator. 
  & 1 & 0 \\

  3 & \scriptsize Excavate once and return to the initial pose; afterward, move to another excavation point and repeat the same sequence once. 
  & 1 & 0 \\

  4 & \scriptsize The dump truck dumps the soil at its current location. 
  & 0 & 1 \\

  5 & \scriptsize The dump truck dumps the soil at the dumping site, then moves to the loading point. 
  & 0 & 1 \\

  6 & \scriptsize Excavate the soil once and transport it to the dumping site using a dump truck. 
  & 1 & 1 \\

  7 & \scriptsize Load soil onto the dump truck, level the soil on its bed, and then move it to the dumping site to unload. 
  & 1 & 1 \\

  8 & \scriptsize Repeat the excavation and dumping sequence twice using the same excavator–dump truck pair. 
  & 1 & 1 \\

  9 & \scriptsize Load soil onto each of the two dump trucks once and transport it. 
  & 1 & 2 \\

  10 & \scriptsize Use two dump trucks to excavate and transport soil a total of four times. Load soil onto each dump truck twice per trip. 
  & 1 & 2 \\

  \bottomrule
  \end{tabularx}
\end{table*}

\subsection{Behavior Tree Generation}

As the second step, Behavior Trees (BTs) are generated by an LLM based on the Action Sequence produced in the previous stage. 
The inputs include the generated Action Sequence, a template structure, BTs example, definitions of custom nodes specific to ROS2-TMS for Construction, and task parameters such as movement paths and joint configurations stored in the system database.

The LLM generates individual BTs for each construction machine, including the excavator and dump truck. 
The generated BTs are formatted in XML according to the specifications of ROS2-TMS for Construction.

\begin{figure}[t]
  \centering
  \includegraphics[width=\columnwidth]{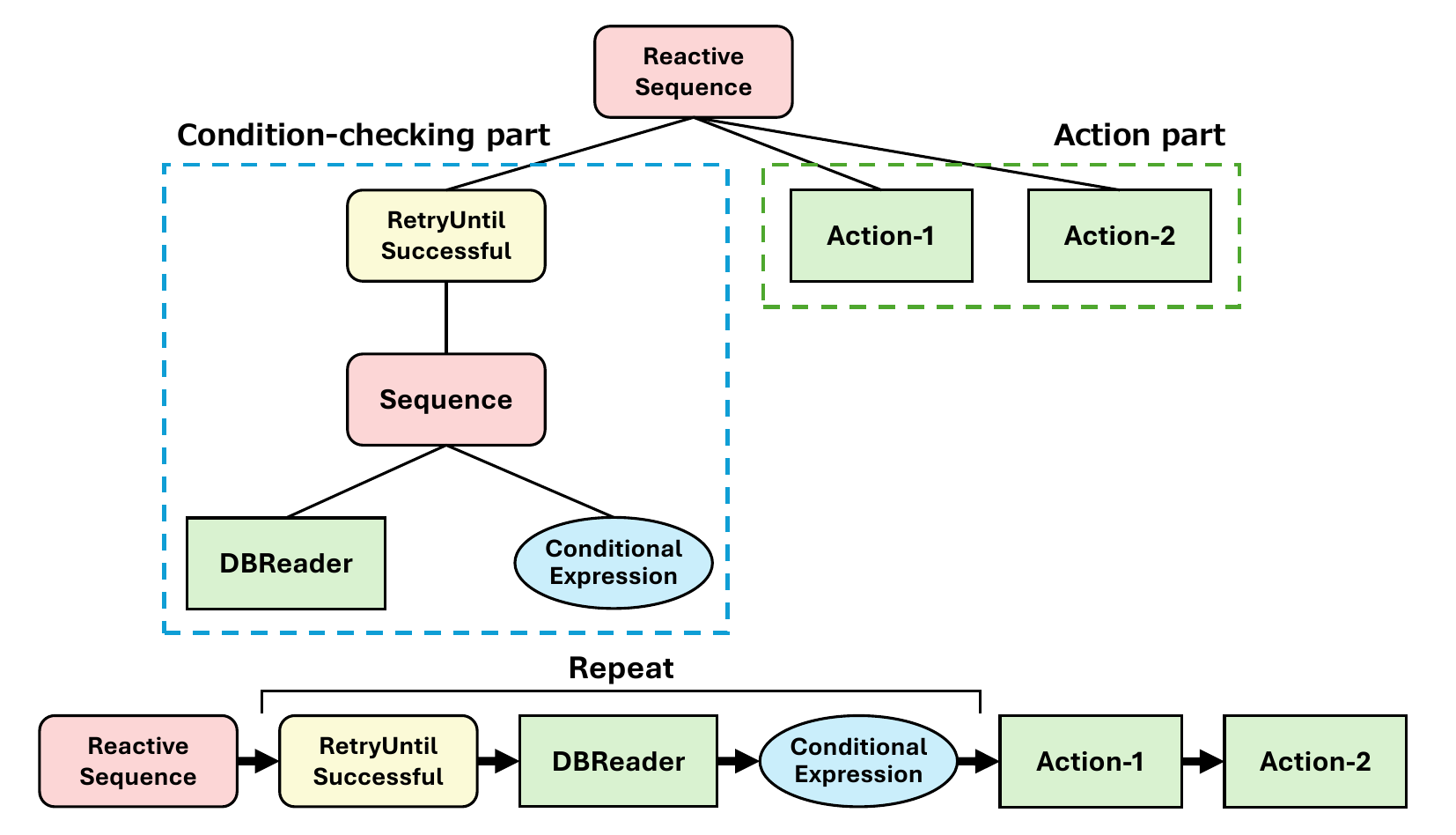}
  \caption{Template structure used for Sequential Behavior Tree generation.
          The DBReader node reads a specified flag from the database (Global Blackboard) into the Local Blackboard, and the ConditionalExpression node evaluates the corresponding precondition.
          The flow at the bottom illustrates tick propagation within the template structure.}
  \label{template_structure}
  \vspace{-1em}
\end{figure}

To guide consistent BTs generation, the structure shown in Fig.~\ref{template_structure} is provided to the LLM as a template. 
Existing approaches for BTs generation using LLMs, such as BETR-XP-LLM~\cite{BETRXPLLM2024} and LLM-as-BT-Planner~\cite{LLMasBTPlanner2025}, employ fallback nodes to extend BTs so that unmet preconditions can be resolved autonomously within a single robot. 
In contrast, in the proposed method, when a construction machine encounters a precondition that cannot be satisfied through its own capabilities, it waits until the required condition becomes satisfied through the operation of another construction machine.

The template structure shown in Fig.~\ref{template_structure} consists of two main components: a condition-checking part and an action part. 
In the condition-checking part, the DBReader node gets flag states corresponding to the \textit{precondition} from the database, and the ConditionalExpression node evaluates the logical condition. 
By repeatedly performing flags reading and condition evaluation through the RetryUntilSuccessful node, the template continuously evaluates flag states until the specified \textit{precondition} becomes satisfied.

Once the \textit{precondition} becomes satisfied, the corresponding action part becomes active. 
If the \textit{precondition} becomes invalid during operation, the ongoing action is interrupted to maintain safe operation. 
To achieve this behavior, a ReactiveSequence node is placed at the root of the template structure. 
Unlike a standard Sequence node, a ReactiveSequence restarts execution from its leftmost child at every tick, allowing the condition-checking part to be continuously evaluated even while actions are active.

The action nodes operate at a low-level granularity, such as moving joints to specified target angles or following predefined paths. 
Complex construction behaviors, including excavation, are expressed by combining these primitive actions. 
The execution of these low-level motions is realized using standard ROS 2 packages such as MoveIt~\cite{moveit} and Navigation2~\cite{nav2}.

The Behavior Tree generation process also supports a HITL. 
Users can review the generated XML-formatted Behavior Trees and provide corrective feedback as additional textual instructions.


\section{EXPERIMENTS AND RESULTS}

\subsection{Simulation Experiments}

\begin{table*}[t]
  \centering
  \vspace{2mm}
  \caption{Experimental results for single and coordinated tasks}
  \label{simulation_experiment_results}

  \begin{tabular}{c cccc @{\hspace{1.5em}} ccccc}
  \toprule
  \multirow{2}{*}{Model}
  & \multicolumn{4}{c}{Single Task}
  & \multicolumn{5}{c}{Coordinated Task} \\

  \cmidrule(lr){2-5}
  \cmidrule(lr){6-10}

  & SR & NN & TU & GT
  & SR & NRF & NN & TU & GT \\
  \midrule

  Claude-Opus-4.6
  & \textbf{1.00} & \textbf{8.92} & 13804.85 & \textbf{10.23}
  & \textbf{0.93} & \textbf{0.27} & \textbf{75.80} & \textbf{20124.93} & \textbf{50.47} \\

  GPT-5.2
  & \textbf{1.00} & 10.40 & \textbf{10749.27} & 16.73
  & \textbf{0.93} & 1.40 & 98.71 & 20905.50 & 143.00 \\

  gpt-oss:120b
  & 0.93 & 10.73 & 13982.67 & 67.73
  & 0.07 & 1.58 & (29.27) & 24787.33 & 367.80 \\

  gpt-oss:120b (w/o HITL)
  & 0.40 & -- & -- & --
  & 0.00 & -- & -- & -- & -- \\

  \bottomrule
  \end{tabular}

\end{table*}

\subsubsection{Dataset}

To evaluate the proposed workflow, we constructed a dataset consisting of 30 natural-language construction instruction scenarios, including 15 single-task scenarios and 15 coordinated-task scenarios.
Table~\ref{scenarios_example} shows some examples of the scenarios. No. 1–5 are single-task scenarios, and No. 6–10 are coordinated-task scenarios.
Single-task scenarios involve operations performed by a single construction machine, such as excavation, gathering, relocation, and dumping, and primarily evaluate instruction interpretation and task decomposition.
Coordinated-task scenarios require cooperation between one excavator and one or two dump trucks, introducing synchronization constraints and shared workspace interactions. 
The scenarios were designed with increasing levels of difficulty by varying task repetitions, spatial transitions, and synchronization dependencies. 
All instruction sentences were designed to reflect realistic earthwork operation requests with linguistic variations.

\subsubsection{Evaluation Metrics}

The performance of the generated plans are evaluated from the following evaluation metrics.

\begin{itemize}

  \item \textbf{SR (Success Rate):}
  The ratio of scenarios in which valid Behavior Trees were successfully generated and executed in simulation without task failure or unsafe machine behavior.

  \item \textbf{NN (Number of Nodes):}
  The average number of nodes contained in the Behavior Trees across all scenarios.
  This metric reflects the structural complexity of the generated plans and the potential presence of redundant nodes introduced during LLM-based generation.

  \item \textbf{NRF (Number of Redundant Flags):}
  The average number of synchronization flags unnecessarily generated, including flags that represent duplicated state conditions or introduce superfluous synchronization within the same machine.

  \item \textbf{TU (Token Usage):}
  The total number of tokens consumed during LLM inference for generating the Action Sequence and Behavior Trees.

  \item \textbf{GT (Generation Time):}
  The total seconds required to generate plans.

\end{itemize}

\subsubsection{Simulation Results}

\begin{figure}[t]
  \centering
  \includegraphics[width=0.8\columnwidth]{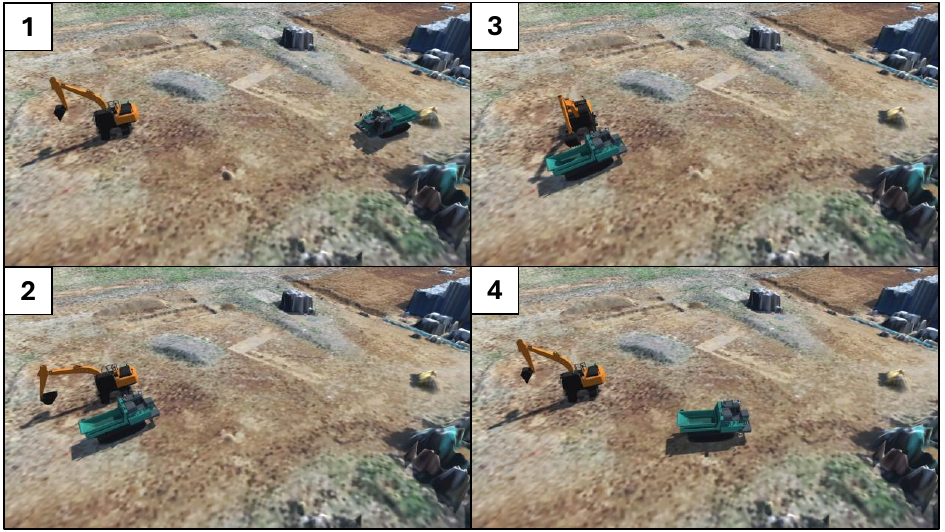}
  \caption{Simulation view of Scenario No.6. The dump truck approaches the excavator (1–2); the excavator excavates soil and loads it onto the dump truck (3); the dump truck moves to the dumping site and dumps soil (4).}
  \label{simulation_experiment_view}
\end{figure}

Experiments were conducted using OperaSim-PhysX~\cite{OperaSim-PhysX}, a Unity-based simulator (Fig.~\ref{simulation_experiment_view}).
For each scenario, Action Sequences and Behavior Trees were generated through the proposed two-stage workflow, allowing at most one HITL refinement per step.
A scenario was considered successful if the generated Behavior Trees executed all planned actions without collision, deadlock, or synchronization violation and fulfilled the given instruction.

As shown in Table~\ref{simulation_experiment_results}, state-of-the-art LLMs, including GPT-5.2 and Claude-Opus-4.6, achieved high success rates in both single-task (100\%) and coordinated-task scenarios (93\%), demonstrating the effectiveness of the proposed two-stage decomposition and synchronization design when paired with capable models.
In contrast, gpt-oss:120b showed significant degradation in coordinated tasks (7\% SR), reflecting the increased difficulty of reasoning about inter-machine synchronization and dependency management, which requires explicit representation of inter-machine dependencies.
Without HITL refinement, it failed in nearly all coordinated scenarios, whereas HITL-enabled generation achieved moderate success in single-task cases.
This result highlights the importance of human feedback in resolving inconsistencies related to synchronization reasoning during plan generation.

\begin{figure}[t]
  \centering
  \includegraphics[width=0.9\columnwidth]{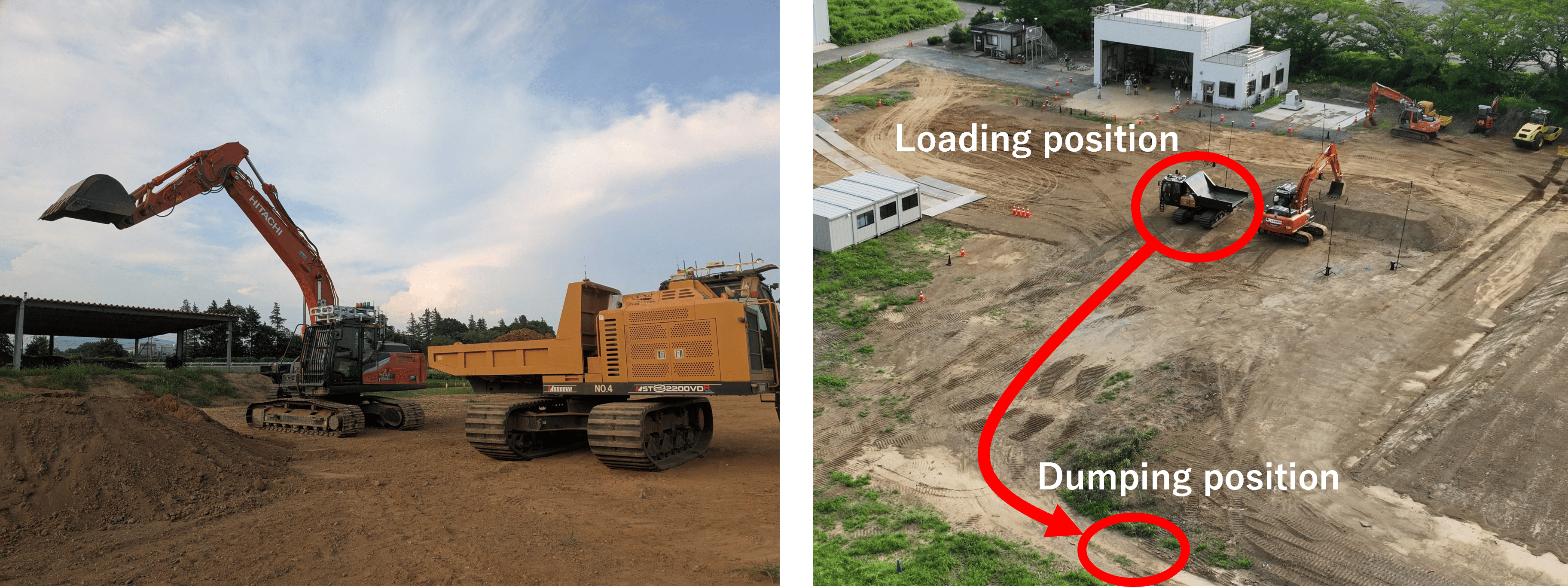}
  \caption{Construction machines and  experimental field. Left: The excavator (ZX200) and the dump truck (MST110CR). Right: The loading position, where the excavator excavates the soil and loads it onto the dump truck, and the dumping position, where the dump truck dumps the soil.}
  \label{real_world_experiment_condition}
  \vspace{-1em}
\end{figure}

The increased planning complexity of coordinated tasks is reflected not only in reduced success rates but also in larger Behavior Trees (NN).
Planning cost also increased in coordinated scenarios, as evidenced by higher token usage (TU) and longer generation time (GT), primarily due to additional reasoning required for synchronization handling and multi-machine task decomposition.
Claude-Opus-4.6 produced more compact trees and shorter generation times than GPT-5.2, particularly in coordinated scenarios.
The low success rate of gpt-oss limits interpretation of its NN values.

Regarding redundant flags (NRF), Claude-Opus-4.6 achieved the lowest value (0.27), while GPT-5.2 and gpt-oss produced higher values (1.40 and 1.58).
The latter models often generated new flags for repeated cycles instead of reusing equivalent ones, increasing redundancy without improving execution validity.
The only failure of Claude-Opus-4.6 occurred in Scenario No.~10 (Table~\ref{scenarios_example}), where a highly complex instruction led to an Action Sequence exceeding 30 lines and prevented simultaneous BT generation due to output token limitations.

\begin{figure*}[t]
  \centering
  \includegraphics[width=\textwidth]{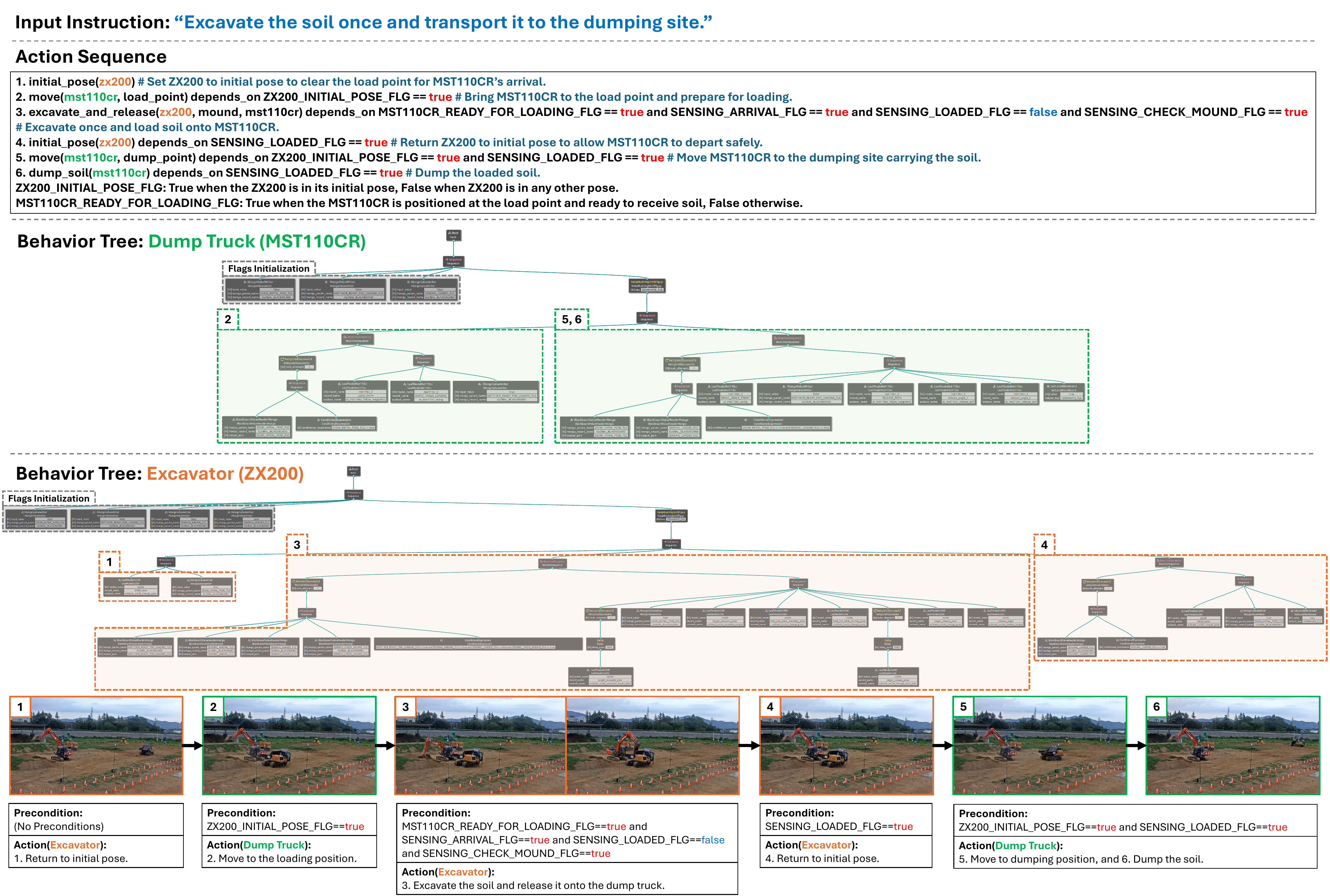}
  \caption{Experimental results of excavation and soil dumping tasks in the real-world. From top to bottom, the figure shows the Action Sequence, the BT for the excavator (ZX200), the BT for the dump truck (MST110CR), and the sequence of operations. The numbering in the high-level task plan (1–6) corresponds to the numbers in the BTs and the operation sequence.}
  \label{real_world_experiments}
  \vspace{-1em}
\end{figure*}

Overall, the results demonstrate that the proposed two-stage workflow with explicit synchronization design enables reliable and scalable coordination of multiple construction machines in simulation when supported by high-capability LLMs.

\subsection{Real-World Experiments}

To verify the practicality of the proposed workflow in real-world environments, we conducted cooperative earthwork experiments using an excavator (ZX200) and a dump truck (MST110CR), as shown in Fig.~\ref{real_world_experiment_condition}.
The input instruction was {\bf ``Excavate the soil once and transport it to the dumping site.''}
Both Action Sequence generation and Behavior Tree generation were performed using GPT-5, which was the state-of-the-art model available at the time the real-world experiments were conducted.

Fig.~\ref{real_world_experiments} illustrates the generated Action Sequence, the corresponding Behavior Trees, and the observed execution process in the experimental field, showing how synchronization between machines emerges during task execution.

At the beginning of execution, the excavator first transitioned to its initial pose while the dump truck remained stationary.
The dump truck waited until the synchronization condition confirming that the excavator had returned to its initial pose was satisfied, allowing the dump truck to safely approach the loading position in the real environment.
Once the excavator reached the initial pose, the corresponding synchronization flag was updated through the Global Blackboard, enabling the dump truck to start moving toward the loading position.
Excavation and loading operations began after sensing information and machine-state conditions were satisfied.
These conditions included dump truck arrival, loading readiness, and soil availability detected by the sensing system.
As a result, the excavator and dump truck operated sequentially without unsafe interference during the cooperative task.

After completing the loading operation, the excavator returned to its initial pose to secure a safe departure path for the dump truck.
Once this synchronization condition was satisfied, the dump truck moved toward the dumping site and successfully discharged the soil.
This sequence confirms that synchronization based on shared flags was consistently maintained throughout multiple stages of the cooperative operation.

The experimental results demonstrate that the proposed method enables synchronization between construction machines through shared flag management, allowing cooperative task execution generated by LLMs to be successfully deployed in a real-world environment.

\section{CONCLUSIONS}

In this study, we proposed an LLM-based workflow for autonomous control of construction machinery by generating Behavior Trees (BTs) within the construction automation platform, ROS2-TMS for Construction.
The proposed method adopts a two-stage generation process consisting of Action Sequence generation for task decomposition and subsequent template-based BTs generation.
It further enables coordinated operation among construction machines through synchronization flags shared via a Global Blackboard.

Simulation experiments demonstrated that state-of-the-art LLMs can generate executable task plans across a wide range of construction instructions, while coordinated tasks increase planning complexity due to synchronization requirements.
The results also highlighted the importance of Human-in-the-Loop correction for improving plan reliability in complex multi-machine scenarios.

Furthermore, real-world experiments using an excavator and a dump truck confirmed that LLM-generated task plans can be successfully applied to real construction machines, enabling safe and coordinated task execution in practical environments.
These results indicate the potential to significantly reduce the manual effort required for BTs design, which has traditionally been a major challenge in deploying coordinated construction machine operations.

Future work will focus on extending the framework to additional construction machine types and improving scalability for large-scale cooperative operations.
In addition, reducing planning cost is an important direction to enable BTs generation using smaller LLMs with fewer parameters, such as open-weight models, while maintaining practical performance comparable to state-of-the-art models.

\section{ACKNOWLEDGMENT}
This work was supported by Council for Science, Technology and Innovation(CSTI), Cross-ministerial Strategic Innovation Promotion Program (SIP), the 3rd period of SIP “Smart Infrastructure Management System” Grant Number JPJ012187 (Funding agency: Public Works Research Institute).

\begingroup
\tiny
\printbibliography
\endgroup

\end{document}